\documentclass[runningheads]{llncs}
\usepackage[T1]{fontenc}

\usepackage{graphicx}

\usepackage{graphicx}
\usepackage{amsmath}
\usepackage{amssymb}
\usepackage{booktabs}
\usepackage{colortbl}
\usepackage{float}
\usepackage{multirow}
\usepackage{array}
\usepackage{placeins}
\usepackage{xcolor}

\usepackage{hyperref}

\usepackage[style=numeric,sorting=none]{biblatex}
\begin{document}
\title{Active few-shot segmentation by reinforcing data selection}
%

\author{Chenlan Zhao\inst{1,\dagger} \and
Benny Wong\inst{1,\dagger} \and
Timothy F. Lundberg\inst{1,\dagger} \and
Ahmed M. Elsayed \inst{1,\dagger} \and
Abdallah Aljarkas \inst{1,\dagger} \and
Hamad A. Aljamaan \inst{1,\dagger} \and
Lynn Karam \inst{1,2,3,4} \and
Qianye Yang \inst{5} \and
Yipeng Hu \inst{3,5} \and
Claire C. Villette \inst{1,2} \and
Shaheer U. Saeed \inst{1,2,3,*}
}
\authorrunning{Zhao et al.}

\institute{Centre for Bioengineering, School of Engineering and Materials Science, Queen Mary University of London, London, United Kingdom
\and
Digital Environment Research Institute, Queen Mary University of London, London, United Kingdom\\
\and
UCL Hawkes Institute; Department of Medical Physics and Biomedical Engineering, University College London, London, United Kingdom\\
\and
Department of Biological and Biomedical Sciences, Yale University, Connecticut, USA\\
\and
Institute of Biomedical Engineering, Department of Engineering Science, University of Oxford, Oxford, United Kingdom\\
$\dagger$Contributed equally to this work\\
*Email: \email{shaheer.saeed@qmul.ac.uk}}

\maketitle

\begin{abstract}
Few-shot learning enables medical image segmentation models to adapt to new tasks using only a small number of labelled examples. However, adaptation performance depends strongly on which examples are selected for the support set. Effective support sets should capture relevant variation within the target domain and be informative for adaptation, with constituent samples providing complementary information. Despite this, existing active data selection approaches largely prioritise samples individually and do not explicitly account for interactions between examples. In this work, we propose a reinforcement learning framework for support-set selection in few-shot medical image segmentation, enabling support sets to be optimised jointly rather than through independent sample scoring. Given a pool of unlabelled candidate images, an agent directly predicts a support set that maximises downstream segmentation performance. Experiments on a cross-institutional pelvic MRI dataset demonstrate improvements over random selection and current state-of-the-art methods. Our findings highlight the importance of support-set complementarity for effective adaptation and demonstrate the potential of reinforcement learning for optimising adaptation sets. 

Code: \url{https://github.com/timfkl/active-few-shot-learning}

\end{abstract}

\section{Introduction}

Medical image segmentation is a fundamental component of many clinical workflows, supporting disease diagnosis, treatment planning, and image-guided interventions \cite{ronneberger2015u, gao2025medical, bhattacharya2022review}. Recent advances in deep learning have substantially improved segmentation performance across a wide range of anatomical structures and imaging modalities \cite{gao2025medical}. However, the success of these approaches is largely driven by access to large quantities of densely annotated training data \cite{peng2021medical}. In medical imaging, many tasks, such as organ segmentation, require dense pixel-level annotations. Producing these is not only time-intensive but requires specialist clinical expertise, while being subject to high inter-observer variability \cite{saeed2024active, czolbe2021segmentation, chalcroft2021development, tajbakhsh2020embracing}. As a result, obtaining sufficient labelled data remains a major challenge, particularly for inherently data-scarce settings such as newly introduced imaging protocols, rare diseases, and under-studied anatomical structures \cite{pachetti2024systematic,saeed2026reasoning}.

Few-shot learning aims to address this limitation by enabling adaptation to new tasks using only a small number of labelled examples. Gradient-based meta-learning methods such as MAML \cite{finn2017model} and Reptile \cite{nichol2018first} learn model initialisations that can be rapidly adapted to unseen tasks from only a handful of support examples. This paradigm has shown considerable promise for medical image segmentation across multiple organs, imaging modalities, and clinical applications under limited annotation budgets \cite{khadka2022meta, alsaleh2024few, li2023prototypical}.

Despite these advances, most few-shot learning methods implicitly assume that support examples are randomly selected or otherwise fixed \cite{nichol2018first}. However, adaptation performance depends not only on the number of support examples available, but also on which examples are selected. In few-shot meta-learning, the support set directly determines the adaptation process and can therefore have a substantial impact on downstream performance \cite{li2023prototypical, tang2025few}. Support sets that provide limited coverage or fail to capture the diversity of the target distribution may lead to poor generalisation. This challenge is particularly relevant in medical imaging, where substantial variation exists in anatomy, pathology appearance, and image quality \cite{renard2020variability, bhattacharya2022review}. As a result, two support sets of identical size may produce markedly different segmentation performance, potentially contributing to the performance gap that often remains relative to fully supervised approaches.

An effective support set should satisfy several desirable properties \cite{wang2024comprehensive, li2024survey}. First, it should aim to provide sufficient coverage of the target distribution to ensure that adaptation is not biased towards a very narrow subset of image appearances or anatomical variations. Second, it should be informative, containing examples that provide strong learning signals for the target task. Third, the selected samples should be complementary, collectively capturing diverse aspects of the target distribution rather than providing redundant information. Existing active learning \cite{saeed2022image, ho2023learning} and data valuation \cite{yoon2020data} methods typically estimate the utility of individual samples before selecting the highest-scoring candidates, or assume larger data volumes \cite{cui2026labeled,kirsch2019batchbald,song2019solving}. However, support-set selection differs fundamentally from these settings because adaptation performance depends on the collective properties of a small selected set. An effective support set must jointly balance coverage, informativeness, and complementarity, making support-set quality a set-level property rather than a sum of individual sample utilities.

In this work, we propose a reinforcement learning framework for few-shot support-set selection. Given a small pool of unlabelled candidate images, an RL agent directly predicts a support set that maximises downstream segmentation performance after adaptation. During training, the selected support set is used to adapt a meta-learned segmentation model, while performance on a validation set provides the reward signal used to optimise the selection policy. By directly optimising post-adaptation generalisation performance, the proposed framework learns support-set selection strategies that balance coverage, informativeness, and complementarity without relying on hand-crafted selection heuristics.

The contributions of this work are threefold: 1) identify support-set selection as an important and underexplored challenge in few-shot medical image segmentation and propose a reinforcement learning framework for support-set optimisation; 2) evaluate the proposed approach on a real-world cross-institutional male pelvic MR segmentation dataset; 3) demonstrate that learned support-set selection improves few-shot segmentation performance compared with conventional random support-set sampling and other state-of-the-art methods.

\section{Methods}

\begin{figure}[!ht]
\centering
\includegraphics[width=0.95\linewidth]{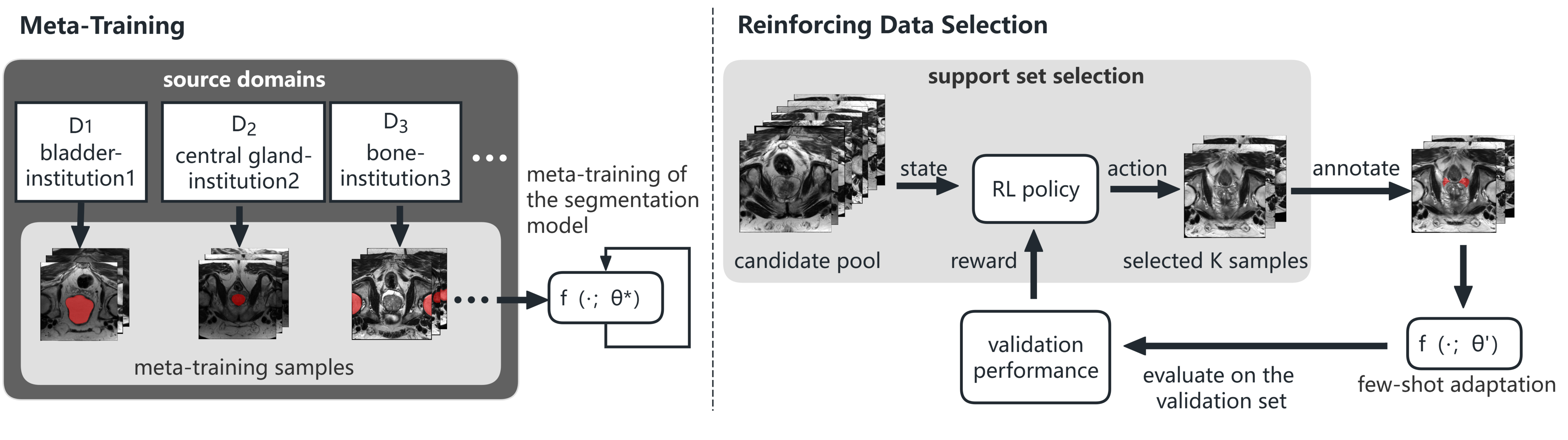}
\caption{Overall framework of the proposed method.}
\label{fig:overall}
\end{figure}

An overview of the proposed framework is shown in Fig.~\ref{fig:overall}. Given a small pool of unlabelled images from a target task, a reinforcement learning (RL) agent selects a support set for few-shot adaptation. The selected images are subsequently labelled and used to adapt a meta-learned segmentation model. Performance on a held-out validation set is then used as a reward signal to optimise the data selection policy. Through this process, the RL agent learns to identify support sets that maximise downstream few-shot segmentation performance.

\subsection{Few-shot learning}

Let $\{\mathcal{D}_i\}_{i=1}^{N}$ denote a collection of source domains, where each domain $\mathcal{D}_i$ defines a distribution over images $x$ and corresponding segmentation labels $y$. A segmentation model $f(\cdot;\theta)$ is meta-trained across the source domains to learn parameters that can be rapidly adapted to previously unseen target domains.

During meta-training, examples $\{(x_{i,j},y_{i,j})\}_{j=1}^{M}$ are sampled from each source domain, with $(x_{i,j},y_{i,j})\sim\mathcal{D}_i$. For simplicity, the meta-training objective can be expressed as minimising the expected segmentation loss across source domains:

\begin{equation}
\theta^* = \arg\min_{\theta}
\mathbb{E}_{\mathcal{D}_i}
\left[
\frac{1}{M}
\sum_{j=1}^{M}
\mathcal{L}
\left(
f(x_{i,j};\theta),y_{i,j}
\right)
\right],
\end{equation}

where $\mathcal{L}$ denotes the segmentation loss and $\theta^{*}$ are the resulting meta-trained parameters. In practice, meta-training was performed using Reptile \cite{nichol2018first}, a first-order meta-learning algorithm that learns model initialisations amenable to rapid adaptation on new domains.

Given a previously unseen target domain $\mathcal{D}_t$, a support set containing $K$ labelled examples, $
\{(x_{t,j},y_{t,j})\}_{j=1}^{K}$, where $
(x_{t,j},y_{t,j})\sim\mathcal{D}_t$, is used to adapt the model. Initialising the model with $\theta^*$, the adapted parameters are obtained according to:

\begin{equation}\label{eq:adaptation}
\theta' = \arg\min_{\theta}
\frac{1}{K}
\sum_{j=1}^{K}
\mathcal{L}
\left(
f(x_{t,j};\theta),y_{t,j}
\right)    
\end{equation}

The adapted model $f(\cdot;\theta')$ is subsequently applied to the remaining images from the target domain.

In this work, the few-shot learner is treated as a fixed adaptation mechanism. Our objective is therefore to identify support sets that maximise downstream adaptation performance.

\subsection{Reinforcement learning for support-set selection}

The objective of the proposed framework is to identify support sets that maximise downstream few-shot adaptation performance. To achieve this, support-set selection is formulated as a reinforcement learning problem.

Given a target domain $\mathcal{D}_t$, a candidate pool containing $P$ unlabelled examples is denoted as $s = \{x_{t,j}\}_{j=1}^{P}$, where $x_{t,j}\sim\mathcal{D}_t$ and the candidate pool forms the state $s\in\mathcal{S}$. The agent then predicts an action $a = \{a_j\}_{j=1}^{P}$, where $a_j \in \{0,1\}$ and $a \in \mathcal{A}=\{0,1\}^P$. This corresponds to a binary selection mask indicating which examples should be included within the support set e.g., $a_j=1$ indicates that the corresponding image is selected for inclusion within the support set. Note that the constraint $\sum_{j=1}^{P} a_j = K$ and $K<P$ applies such that exactly $K$ examples are selected in the support set. The selected support set is subsequently labelled and used to adapt the few-shot segmentation model as described in equation \ref{eq:adaptation}.

Following adaptation, the model is evaluated on a validation set from the target domain $\mathcal{D}_t$ containing $V$ labelled examples, $\{(x_{t,j},y_{t,j})\}_{j=1}^{V}$, where $(x_{t,j},y_{t,j})\sim\mathcal{D}_t$. The reward is formulated as follows:

\begin{equation}
r
=
\frac{1}{V}
\sum_{j=1}^{V}
\mathrm{Dice}
\left(
f(x_{t,j};\theta'),
y_{t,j}
\right),
\end{equation}

where $\theta'$ denotes the adapted model parameters.

The agent is parameterised by $\phi$ and learns a policy $\pi(\cdot;\phi)$ that maps candidate image pools to support-set selections $a = \pi(s;\phi)$. For example, given a candidate pool containing $P=8$ images and a support-set size of $K=4$, the agent may predict $a = [1,0,1,1,0,0,1,0]$ corresponding to the selection of images $\{x_{t,1},x_{t,3},x_{t,4},x_{t,7}\}$ for inclusion within the support set.

The policy parameters are optimised to maximise the expected reward: 

\begin{equation}
   \phi^* =  \arg\max_{\phi}
\mathbb{E}_{s,a} \left[r\right]
\end{equation}

Importantly, the reward is only observed after support-set selection, annotation, and few-shot adaptation, encouraging the policy to identify support sets that maximise downstream generalisation rather than individual sample utility. In practice, policy optimisation was performed using proximal policy optimisation (PPO) \cite{schulman2017proximal}. By directly optimising validation performance following adaptation, the proposed framework learns support-set selection strategies that jointly account for coverage, informativeness, and complementarity without relying on hand-crafted selection heuristics.

\subsection{Inference}

During inference, a previously unseen target domain $\mathcal{D}_t$ is provided together with a candidate pool of $P$ unlabelled examples. The trained policy $\pi(\cdot;\phi^*)$ predicts a support-set selection mask $a$, which is used to select $K$ examples for annotation. The resulting support set of $K$ samples is then annotated and used to adapt the meta-trained segmentation model according to Eq.~\ref{eq:adaptation}, yielding adapted parameters $\theta'$. The adapted model is subsequently applied to the remaining images from the target domain. No policy optimisation or reward computation is performed during inference.

\section{Experiments}

\subsection{Dataset}
Experiments were conducted using the cross-institutional 3D male pelvic MRI dataset introduced by Li et al. \cite{li2023prototypical}. The dataset comprises 589 T2-weighted 3D MR images acquired across eight institutions and includes annotations for eight pelvic anatomical structures: bladder, bone, obturator internus, transition zone, central gland, rectum, seminal vesicle, and neurovascular bundle. Prior to training, all images and segmentation masks were resampled and resized to dimensions $256\times256\times32$.

To evaluate few-shot adaptation under realistic data-scarce conditions, the neurovascular bundle (NB) and obturator internus (OI) were reserved exclusively for evaluation, while the remaining structures were used during meta-training. These structures were selected due to their relatively limited availability in routine clinical annotation workflows, making them representative of practical few-shot segmentation scenarios. In addition, institutions 3 and 4 were held out entirely during training to provide an unseen target-domain evaluation setting, while the remaining institutions were used for meta-training. Evaluation was therefore performed on NB and OI cases originating from institutions 3 and 4, comprising 156 volumes (74 and 82, respectively). For each evaluation experiment, up to 16 labelled examples were used for support-set adaptation and a further 8 examples were used as the validation set for reinforcement learning reward computation. Remaining volumes were reserved for evaluation.

\subsection{Network architectures and hyper-parameters}

All backbone networks in experiments followed a UNet architecture consisting of three downsampling stages in the contracting path and three corresponding upsampling stages in the expanding path. The network receives a grayscale image as input (single channel) and initializes with a base feature depth of 8 channels. The number of feature channels doubles at each encoder stage through successive $3 \times 3 \times 3$ convolutional layers followed by Group Normalization, resulting in a 64-channel feature representation at the bottleneck layer. A hybrid binary cross-entropy and Dice loss is used for training. The Reptile algorithm \cite{nichol2018first}, was used for meta-training and adaptation. 

The meta-training took approximately 24 hours on two Nvidia A5000 GPUs (over 10,000 steps) and the RL training took approximately 72 hours on the same hardware. Example code and further hyper-parameter details are provided at \url{https://github.com/timfkl/active-few-shot-learning}.

\subsection{Comparisons}

The proposed support-set selection policy was compared against three alternative selection strategies: (1) random selection, representing the conventional few-shot adaptation setting where support examples are chosen without prioritisation; (2) task-based prioritisation (TBP) \cite{saeed2024active, saeed2022image, saeed2021adaptable, ho2023learning}, an active learning approach that prioritises individual samples according to their estimated utility for the target task; and (3) Data Valuation using Reinforcement Learning (DVRL) \cite{yoon2020data}, which assigns utility scores to individual samples using RL. 

Comparisons were performed using support-set sizes of $K\in\{4,8,16\}$ examples, where $P=32$. Note that in actuality $K$ remained fixed at $K=4$ and where $K>4$, selection was repeated multiple times with randomly sampled candidate pools. For each setting, the selected support examples were used to adapt the Reptile segmentation model and performance was evaluated on the held-out test set using the Dice similarity coefficient. Results are reported separately for NB and OI, as well as combined across both structures. 

Two ablation studies were additionally conducted. First, candidate pool size was varied, $P\in\{8,16,32\}$, while fixing the support-set size to $K=4$ in order to evaluate the impact of the available selection pool. Second, MAML \cite{finn2017model} was evaluated as an alternative few-shot learning backbone to assess generalisation across meta-learning algorithms.

Statistical significance was assessed using paired two-sided t-tests between the proposed method and each comparison approach, with pairing performed at the patient level on the held-out test set.

\section{Results}

The results are summarised in Table~\ref{tab:main_results}. The proposed support-set selection strategy consistently achieved the highest Dice scores across all support-set sizes and both anatomical structures ,with statistical significance (all p-values$<$0.01). Improvements were particularly pronounced in the low-data regime, where support-set quality is expected to have the greatest impact on adaptation performance. For example, at $K=4$, the proposed method showed an 8.1 percentage point increase compared to random selection, and a 4.5 percentage point increase compared to DVRL (statistical significance was not found for the comparison between DVRL and TBP, p-values$>$0.08). Performance improved for all methods as the number of support examples increased, reflecting the availability of additional adaptation data. However, the proposed method maintained a consistent advantage over competing approaches across all support-set sizes, suggesting that the learned policy identifies support sets that are more informative for downstream adaptation, while providing complementary information. Notably, improvements were observed over both TBP and DVRL, which prioritise samples based on individual utility estimates, supporting the hypothesis that effective support-set construction depends on the collective properties of the selected examples rather than the utility of individual samples alone. All observed improvements were statistically significant ($p<0.01$).

\begin{table*}[!ht]
\centering
\caption{Comparison of support-set selection strategies for few-shot segmentation. Results are reported as Dice score (\%, mean $\pm$ standard deviation) on the held-out test set. Statistical significance is assessed against the proposed method using paired t-tests where bold indicates p-value$<$0.01 for all comparisons in column.}
\label{tab:main_results}
\resizebox{\textwidth}{!}{
\begin{tabular}{lccccccccc}
\toprule
&
\multicolumn{3}{c}{$K=4$} &
\multicolumn{3}{c}{$K=8$} &
\multicolumn{3}{c}{$K=16$} \\
\cmidrule(lr){2-4}
\cmidrule(lr){5-7}
\cmidrule(lr){8-10}
Method
& NB & OI & Combined
& NB & OI & Combined
& NB & OI & Combined \\
\midrule

Random
& $48.5\pm6.5$ & $53.8\pm6.2$ & $51.2\pm6.5$ & $55.7\pm5.9$ & $60.1\pm5.6$ & $57.9\pm5.8$ & $61.9\pm5.1$ & $65.8\pm5.0$ & $63.9\pm5.1$ \\

TBP
& $51.6\pm6.0$ & $56.0\pm5.8$ & $53.8\pm5.8$ & $58.8\pm5.2$ & $62.9\pm5.2$ & $60.9\pm5.3$ & $64.3\pm4.8$ & $67.6\pm4.6$ & $66.0\pm4.7$ \\

DVRL
& $52.4\pm5.9$ & $57.1\pm5.6$ & $54.8\pm5.7$ & $59.6\pm5.3$ & $63.7\pm5.3$ & $61.7\pm5.2$ & $65.1\pm4.8$ & $68.4\pm4.5$ & $66.8\pm4.6$ \\

\textbf{Proposed}
& $\mathbf{56.9\pm5.7}$ & $\mathbf{61.7\pm5.9}$ & $\mathbf{59.4\pm5.8}$ & $\mathbf{63.1\pm4.7}$ & $\mathbf{67.9\pm4.6}$ & $\mathbf{65.5\pm4.6}$ & $\mathbf{68.7\pm4.2}$ & $\mathbf{72.4\pm4.1}$ & $\mathbf{70.6\pm4.3}$\\

\bottomrule
\end{tabular}
}
\end{table*}

\noindent\textbf{Ablations:}
We conducted two ablation studies to investigate the behaviour and generalisability of the proposed support-set selection framework. First, we varied the candidate pool size while fixing the support-set size to $K=4$. Increasing the candidate pool size from $P=8$ to $P=16$ and $P=32$ improved the combined Dice score from $56.5\pm5.4$\% to $58.3\pm5.4$\% and $59.4\pm5.8$\%, respectively. This suggests that larger candidate pools provide greater opportunity for identifying informative and complementary support examples, leading to improved downstream adaptation performance. While performance improved between $P=16$ to $P=32$, statistical significance was not observed for the comparison (p-value=0.09). Increasing the candidate pool beyond $P=32$ is computationally expensive, beyond our compute budgets. 

Second, we evaluated the proposed support-set selection strategy using MAML \cite{finn2017model} in place of Reptile. For MAML we observed Dice score $57.1\pm5.9$\% and for Reptile this was $59.4\pm5.8$\%, for $P=32$, $K=4$. Statistical significance was not observed for this comparison (p-value=0.06). This indicates that the benefits of the proposed approach are not specific to Reptile and generalise across different gradient-based meta-learning frameworks.

\begin{figure}
    \centering
    \includegraphics[width=0.8\linewidth]{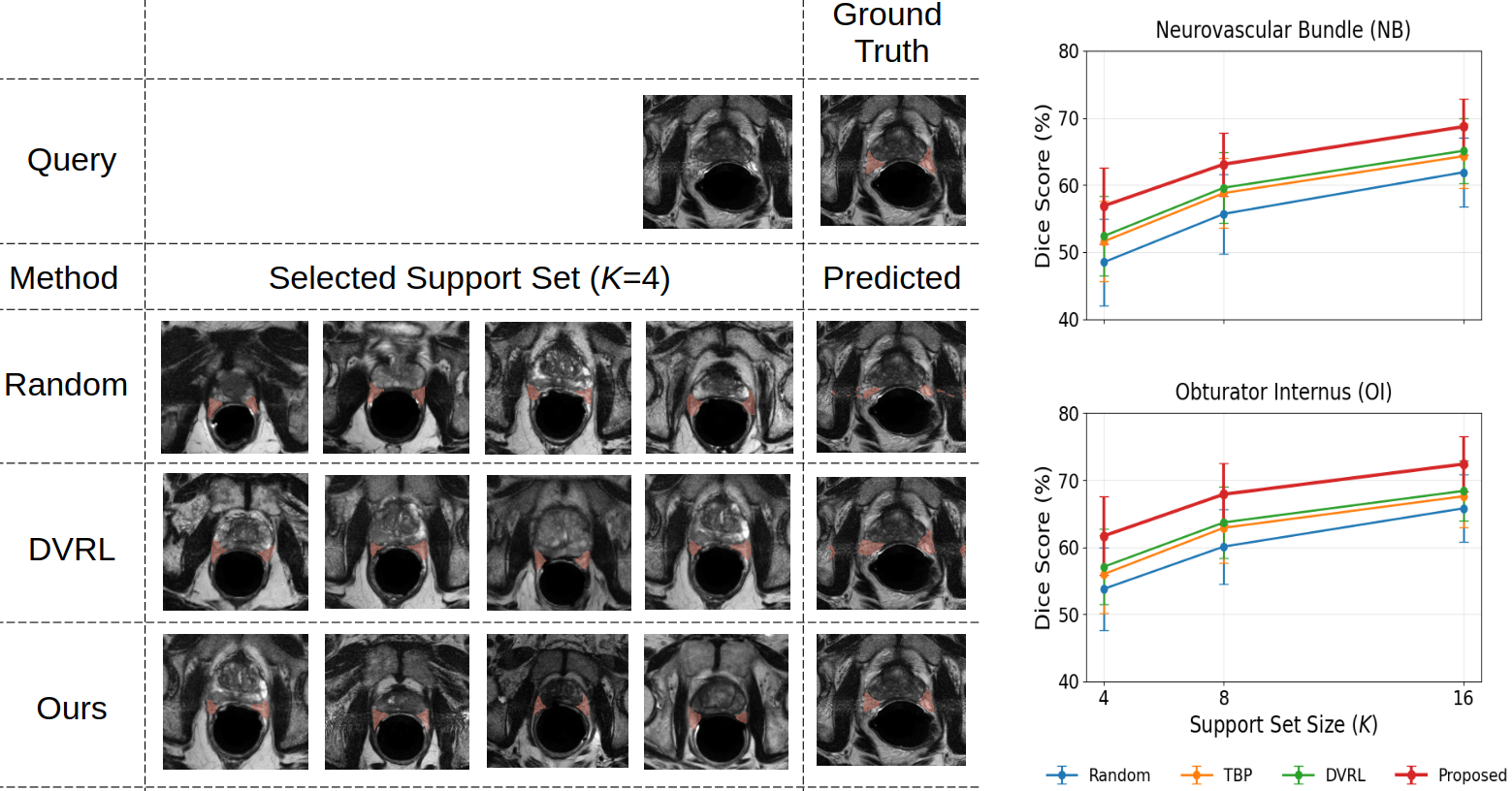}
    \caption{Qualitative results from our proposed method, showing selected support sets.}
    \label{fig:results_fig}
\end{figure}

\noindent\textbf{Qualitative analysis:}
Qualitative results are presented in Fig.~\ref{fig:results_fig}. The query example contains a distortion artefact in the centre of the image, a characteristic present in approximately 10\% of cases within the dataset and associated with a specific scanner at institution 3. Interestingly, the proposed method selected a distorted image within the support set, whereas none of the comparison methods did. Furthermore, the support examples selected by the proposed method exhibited a broader range of image appearances and intensity distributions, while DVRL selected several visually similar examples due to its sample-wise selection strategy. As a result, the proposed method more accurately segmented the distorted region, whereas both Random and DVRL exhibited oversegmentation in areas affected by the artefact. These findings suggest that support-set complementarity is important for few-shot adaptation, enabling the model to better capture variations present within the target domain.

\section{Conclusion}

We presented a RL framework for support-set selection in few-shot medical image segmentation. By directly optimising downstream adaptation performance, the proposed approach learns support sets that balance coverage, informativeness, and complementarity, rather than prioritising samples based solely on individual utility. Experiments on a cross-institutional pelvic MRI dataset demonstrated consistent improvements over random selection, TBP, and DVRL across multiple support-set sizes and anatomical structures. These findings identify support-set selection as an important and underexplored component of few-shot learning and suggest that explicitly modelling interactions between support examples can substantially improve adaptation performance.

\section*{Acknowledgements}

This work is supported by the International Alliance for Cancer Early Detection, an alliance between Cancer Research UK [EDDAPA-2024/100014] \&  [C73666/A31378], Canary Center at Stanford University, the University of Cambridge, OHSU Knight Cancer Institute, University College London and the University of Manchester.

\newpage

\printbibliography

\end{document}